\documentclass{article}

\PassOptionsToPackage{numbers, compress}{natbib}
\usepackage[final]{mlsafety_neurips_2022}

\usepackage[utf8]{inputenc} % allow utf-8 input
\usepackage[T1]{fontenc}    % use 8-bit T1 fonts
\usepackage{hyperref}       % hyperlinks
\usepackage{url}            % simple URL typesetting
\usepackage{booktabs}       % professional-quality tables
\usepackage{amsfonts}       % blackboard math symbols
\usepackage{nicefrac}       % compact symbols for 1/2, etc.
\usepackage{microtype}      % microtypography
\usepackage{xcolor}         % colors

\usepackage{natbib} % has a nice set of citation styles and commands
\usepackage{mathtools} % amsmath with fixes and additions
\usepackage{siunitx} % for proper typesetting of numbers and units
\usepackage{tikz} % nice language for creating drawings and diagrams
\usepackage{algorithm}
\usepackage{algpseudocode}
\usepackage{caption}
\usepackage{subcaption}
\usepackage{bm}
\usepackage{bbm}
\usepackage{enumitem}
\usepackage{wrapfig}
\usepackage{enumitem}
\setlist{leftmargin=5.5mm}

\title{Policy Resilience to Environment Poisoning Attacks on Reinforcement Learning}

\author{
    Hang Xu \\
    School of Computer Science and Engineering \\
    Nanyang Technological University, Singapore \\
    \texttt{hang017@e.ntu.edu.sg}
    \And
    Xinghua Qu \\
    ByteDance AI Lab \\
    Singapore \\
    \texttt{quxinghua17@gmail.com} \\
    \And
    Zinovi Rabinovich \\
    School of Computer Science and Engineering \\
    Nanyang Technological University, Singapore \\ 
    \texttt{zinovi@ntu.edu.sg} 
}

\begin{document}
\maketitle

\begin{abstract}

This paper investigates policy resilience to training-environment poisoning attacks on reinforcement learning (RL) policies, with the goal of recovering the deployment performance of a poisoned RL policy.
Due to the fact that the policy resilience is an add-on concern to RL algorithms, it should be resource-efficient, time-conserving, and widely applicable without compromising the performance of RL algorithms.
This paper proposes such a policy-resilience mechanism based on an idea of knowledge sharing. 
We summarize the policy resilience as three stages: preparation, diagnosis, recovery. 
Specifically, we design the mechanism as a federated architecture coupled with a meta-learning manner, pursuing an efficient extraction and sharing of the environment knowledge. 
With the shared knowledge, a poisoned agent can quickly identify the deployment condition and accordingly recover its policy performance.
We empirically evaluate the resilience mechanism for both model-based and model-free RL algorithms, showing its effectiveness and efficiency in restoring the deployment performance of a poisoned policy. 

% When armed with shared knowledge, a poisoned agent can effectively identify the dynamics model of the deployment environment from limited interactions.
% In this way, imagined trajectories can be derived from the dynamics model, which is then used to recover the deployment performance of the poisoned policy.
% We summarize such a policy resilience as three stages, namely preparation, diagnosis, and recovery.
% Finally, we empirically evaluate the policy-resilience mechanism for both model-based and model-free agents, showing successful and efficient policy recovery. 

\end{abstract}

%%%%%%%%%%%%%%%%%%%%%%%%%%%%%%%%%%%%%%%%%%%
\section{Introduction} %%%%%%%%%%%%%%%%%%%%
%%%%%%%%%%%%%%%%%%%%%%%%%%%%%%%%%%%%%%%%%%%

%% 动机：研究 environment poisoning attack 的解决方案的必要性
The security of Reinforcement Learning (RL) becomes increasingly significant as RL systems have been widely adopted in real-world applications \cite{pan2017virtual,sallab2017deep,kiran2021deep, kim2018reinforcement,sogabe2018smart,lee2020federated, eghbali2021patient,coronato2020reinforcement,riachi2021challenges}.
Thus, it is essential to protect RL-based applications against a variety of adversarial attacks \cite{ma2019policy, zhang2020adaptive, huang2017adversarial, lin2017detecting, rakhsha2020policy}.
Among these attacks, environment poisoning attacks (EPAs) \cite{xu2021transferable, xu2022spiking} are considered particularly insidious.
They poison RL policies by perturbing the training-environment {\em causal factors} (i.e., {\em hyper-parameters}), such as surface frictions, which are exposed outside of RL agents and easily accessible to third parties.
%% 本章要解决的问题
In this paper, we attempt to address the security threat of EPAs from the perspective of resilience. 
The resilience, in this context, refers to an RL agent's ability to recover its policy from malicious manipulations \cite{behzadan2017whatever}.

%% 引出 “分享环境” 的中心思想
In existing EPAs \cite{xu2021transferable, xu2022spiking}, the dynamics of training environment are manipulated as a function of hyper-parameters. 
In other words, the poisoned environment retains the same environment structure (e.g., function) as the natural one, but differs in hyper-parameters. 
%% 完善逻辑链条 - 
As a result, if an RL agent is equipped with the knowledge of environment structures and the capability of identifying hyper-parameters, it will be able to grasp the environment dynamics more efficiently than an agent that learns from scratch.
% based on a large number of interaction samples. 
%% 
% We therefore ask the following questions: can we design a resource-efficient and time-conserving policy-resilience mechanism by exploiting the knowledge of training-environment structure?
% This paper provides a solution to this question.
Accordingly, this paper designs a resource- and time-efficient policy resilience by exploiting the knowledge of environment structure.
%and the capability of \xh{quick penalization}. 

%% 主要思想 
% In this paper, we propose a policy-resilience mechanism in which the poisoned RL agent exploits the environment-structure knowledge to quickly grasp the dynamics of the deployment environment and accordingly recover its policy for optimal deployment performance.
In this paper, the proposed policy resilience can be summarized as three stages: preparation, diagnosis and recovery.
First, the preparation stage intends to extract the critical knowledge of environment structures during the training of RL policies. 
This is achieved based on a design of federated framework incorporating with a meta-learning approach.  
Then, the diagnosis stage occurs prior to the deployment of the learned policy. 
At this stage, the RL agent is shared with the environment knowledge so that it efficiently identifies the dynamics of its deployment environment. 
At the recovery stage, the agent imagines trajectories based on its accurate understanding of the environment dynamics. Accordingly it recovers the poisoned policy for an optimal deployment performance.
% According to an accurate understanding of deployment environment dynamics, the agent can recover its poisoned policy using imagined trajectories and then present an optimal behavior during deployment. 
\begin{wrapfigure}{r}{0.55\textwidth}
    \begin{minipage}{0.55\textwidth}
        \vspace{-0.8cm}
            \begin{figure}[H]
                \centering
                \includegraphics[width=\linewidth]{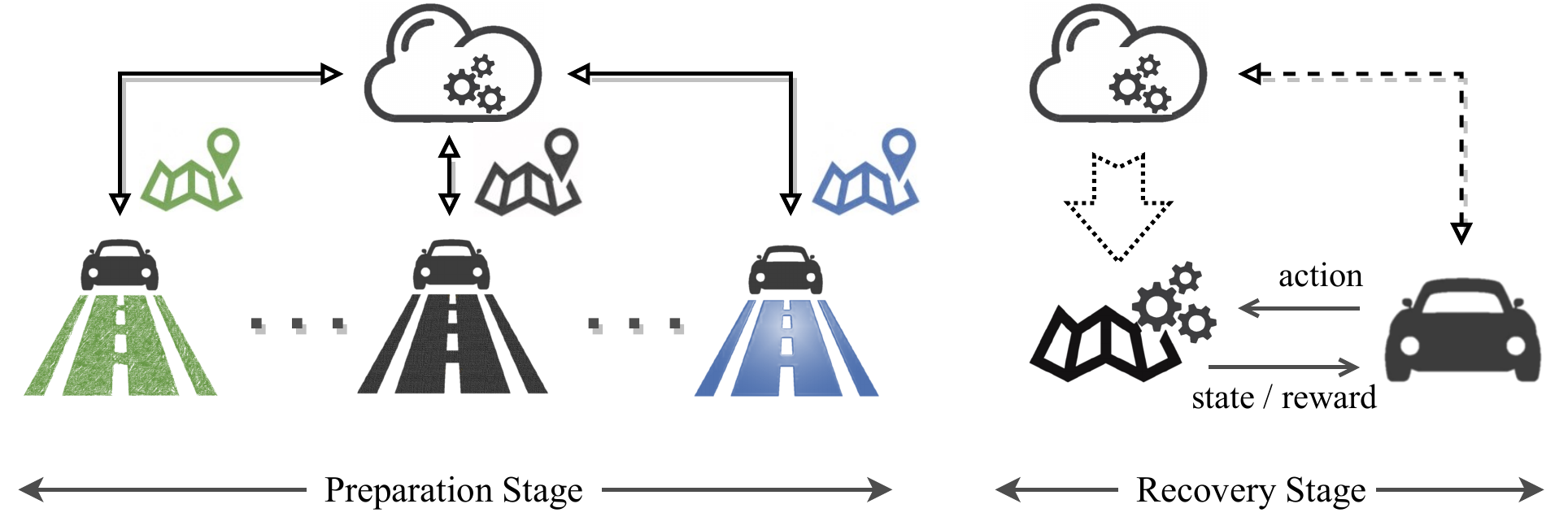}
                \caption{Example of policy-resilience mechanism.}
                \label{fig:federated_illustration}
            \end{figure}
        \vspace{-0.8cm}
    \end{minipage}
\end{wrapfigure}
%% to 节能省力, 引出 federated 设计
Furthermore, it is important to note that the proposed mechanism organizes independent and isolated RL systems in a federated manner, which enables the extraction and sharing of environment knowledge.
% Within each of these RL systems, the environment dynamics share a common underlying structure, which are represented as a parameterized function of hidden parameters (i.e., hyper-parameters). These can be formally defined as an instance of Hidden Parameter Markov Decision Process (Hip-MDP) \cite{killian2017robust}. 
% As a result of these RL systems being organized in a federated manner, 
% As a result, their common environment-structure knowledge can be extracted and fused in the center server, and then shared among federated RL systems.
%% 举例
The federated design is common in RL-based applications.
For example, autonomous vehicles could be organized by a transportation agency in a federated manner, as shown in Figure \ref{fig:federated_illustration}.
The transportation agency (i.e., server) can collect road conditions (e.g., client information) from multiple vehicles and share the merged information to each. In this way, it could assist a single vehicle to drive well in different environment instances. 
Similar examples include robots connected via a cloud server, buildings organized by a smart city center, and financial companies supervised by a central bank. 
These examples imply our proposed policy-resilience mechanism can be applicable to many real-world scenarios. 

As a summary, this paper makes the following contributions:
\begin{itemize}
    \item We propose a policy-resilience mechanism specifically for environment poisoning attacks \cite{xu2021transferable,xu2022spiking} on RL. In this mechanism, independent RL systems are organized into federated systems, which allows each of them to exploit shared knowledge to facilitate the identification of deployment environments and the recovery of individual policies.
    \item The policy resilience can be described as a three-step procedure that includes preparation, diagnosis and recovery. Meta-learning approach is incorporated along with federated manner in the stage preparation and diagnosis, to ensure an efficient extraction of environment knowledge and an accurate interpretation of deployment dynamics.
    \item The policy-resilience mechanism has been evaluated empirically on both model-free and model-based RL agents, which demonstrate that poisoned policies can be recovered in a resource-efficient and time-efficient manner.
\end{itemize}

%%%%%%%%%%%%%%%%%%%%%%%%%%%%%%%%%%%%%%%%%
\section{Problem Statement}  %%%%%%%%%%%%
%%%%%%%%%%%%%%%%%%%%%%%%%%%%%%%%%%%%%%%%%

\paragraph{Description of Environment Poisoning Attacks:}
Our policy-resilience mechanism targets the environment-poisoning attack \cite{xu2021transferable, xu2022spiking} against RL at training-time. 
The attack objective is to force an RL agent (i.e., victim) to learn an attacker-desired policy via minimized changes to the training environment.
% minimized changes to the agent's training environment.
%
% The task of poisoning a victim's policy is performed by an attacker that operates on a different timescale from the victim. 
Specifically, at intervals of the victim's learning process, the attacker manipulates the victim's environment hyper-parameters responding to the victim's behaviour policy and training-environment conditions, until the victim obtains the policy designed by the attacker.

\paragraph{Formulation of Policy Resilience:}

%% HiP-MDP
Our policy-resilience mechanism organizes multiple independent and isolated RL systems in a federated manner.
% Within each of these RL systems, the environment should share a common underlying structure and could differ in hyper-parameters. 
% As a result, these federated RL systems can be formally defined as instances of the same Hidden Parameter Markov Decision Process (Hip-MDP) \cite{killian2017robust} that represents a family of MDPs where hyper-parameters $e \in \mathbb{R}^n$ are used to parameterize the transition dynamics.
As shown in Figure \ref{fig:federated_illustration}, the federated policy-resilience framework consists of one server and a set of client RL systems, each with learnable dynamics models.
We formalize the framework as $\langle \mathcal{S}, \{\mathcal{C}_i\}_{i=1}^{K} \rangle$. $\mathcal{S}$ represents a dynamics model in the reliable server, which is a function parameterized by $\theta_{\mathcal{S}}$. 
$\{\mathcal{C}_i\}_{i=1}^{K}$ represents the set of client RL systems governed by the server, where $K$ is the number of systems. 
Within each of these RL systems, the environment should share a common underlying structure and could differ in hyper-parameters. 
Specifically, each client system is described as a tuple $\mathcal{C}_i = \langle \mathcal{R}_i, \mathcal{M}_i, \mathcal{D}_i, \mathcal{T}_i \rangle$.
Here, $\mathcal{R}_i$ is an RL agent that seeks to learn a policy $\pi$ to maximize cumulative rewards using either model-free or model-based learning algorithms.
$\mathcal{M}_i$ denotes a Markovian environment $\langle s, a, F_{e_i}, r, \gamma \rangle$ (refer to Appendix \ref{appendix-preliminaries}). 
Note that $\{\mathcal{M}_i\}_{i=1}^{K}$ belong to the same Hidden Parameter Markov Decision Process (Hip-MDP) \cite{killian2017robust} with hyper-parameters $e_i\sim p(e)$ which parameterize the dynamics function $F_{e_i}$.
$\mathcal{D}_i$ is a local dynamics model that shares the same structure as the server one $\mathcal{S}$.
And $\mathcal{T}_i$ is a buffer that stores the RL agent's transitions $\langle s, a, s' \rangle$ used for learning a local dynamics model $\mathcal{D}_i$.

Importantly, we assume that the security of the federated framework is guaranteed. This assumption allows us to focus on the security threats occurring within a single isolated RL system. 
Additionally, it is forbidden for each RL agent to directly interact with other agents' environments or access raw transitions in others' buffers.
Instead, clients share local information with the server and trust the knowledge retrieved from the server.

% %% --- figure: preparation & framework ---
% \begin{figure*}[ht]
%     \centering
%     \includegraphics[width=0.9\linewidth]{Figures/federated_framework_3.pdf}
%     \caption{Illustration of preparation stage in policy-resilience mechanism.}
%     \label{fig:resilience_framework}
% \end{figure*}

% %% ----- figures: examples 
% \begin{figure*}[ht]
%     \centering
%     \includegraphics[width=0.8\linewidth]{Figures/pipeline_illustration.pdf}
%     \caption{An example of policy-resilience mechanism in RL systems.}
%     \label{fig:FRL_examples}
% \end{figure*}

% \begin{wrapfigure}{r}{0.41\textwidth}
%     \begin{minipage}{0.41\textwidth}
%         \vspace{-0.8cm}
%             \begin{figure}[H]
%                 \centering
%                 \includegraphics[width=0.8\linewidth]{Figures/pipeline_illustration.pdf}
%                 \caption{3D grid world with size 6x6}
%                 \label{fig:3D_grid}
%             \end{figure}
%         \vspace{-0.8cm}
%     \end{minipage}
% \end{wrapfigure}

%%%%%%%%%%%%%%%%%%%%%%%%%%%%%%%%%%%%%%%%%%%%%%%%%%%%%%%%%%%%%%%
\section{Methodology}  %%%%%%%%%%%%
%%%%%%%%%%%%%%%%%%%%%%%%%%%%%%%%%%%%%%%%%%%%%%%%%%%%%%%%%%%%%%%

%% 详细的介绍 pipeline 

% Based on the design of federated framework, our proposed policy resilience involves three stages, namely, preparation, diagnosis, and recovery.
%
In this section, we introduce our policy-resilience mechanism in accordance with the procedure, namely, preparation, diagnosis and recovery. 
First, we describe mathematically how the environment knowledge is acquired using a meta-learning approach within the federated framework (i.e., the preparation stage).
After that, we describe how the poisoned policy can be recovered in response to an accurate understanding of the deployment environment (i.e., the diagnosis and recovery stages).  

%% ----- preparation -----
\paragraph{Preparation:}

Preparation occurs during the training of RL policies, aiming to efficiently extract critical knowledge about the environment structure.
% which is approximated using the server dynamics model.
We utilize a meta-learning approach \cite{finn17maml} to learn the server dynamics model $\mathcal{S}$ based on environment information collected from all the federated RL systems.
% 使用 meta 的原因
Here, $\mathcal{S}$ is learned as a parameterized function, which can be used to quickly model environments that share a common structure but differ in hyper-parameters.
% 解释使用 meta-learning 的原因 = 突出 meta-learning = learn to learn 的特点
As shown in Figure \ref{fig:resilience_framework} in Appendix \ref{appendix-methodology}, the learning of $\mathcal{S}$ consists of four steps: 
(1) the transfer of dynamics-model parameters (from the server to clients);
(2) the calculation of loss value (in each client system); 
(3) the collection of loss values (from clients to the server); 
(4) the updating of dynamics-model parameters (in the server).
Such a process whereby the server transmits information to client systems and then receives messages from them is regarded as a round. 
% As we follow these four steps at each round, we learn the server dynamics model $\mathcal{S}$ through a combination of federated learning and meta-learning approaches \cite{finn17maml}. 
% Here, with the use of the meta-learning approach, the server dynamics model $\mathcal{S}$ is learned as a parameterized function, which can be easily adapted to environments that share a common structure but differ in hyper-parameters.

%%%----------------- meta-learning + federated learning ---------
% step-1
Mathematically, at the first step of each round, local dynamics models $\{ \mathcal{D}_{i} \}_{i=1}^{K}$ should be initialized with the parameters $\theta_{\mathcal{S}}$ which are retrieved from the server one $\mathcal{S}$.
% step-2
After the initialization, loss values are locally calculated in a meta-learning way \cite{finn17maml}. 
In more detail, in each client RL system where the environment is parameterized by hyper-parameters $e_i$, transitions $\langle s,a,s'\rangle$ are collected during the learning of RL agent's policy. 
These transitions are saved in the local buffer $\mathcal{T}_i$, which is separated into a support set $\mathcal{T}_{e_i}^{spt} = \left \{ \langle s_m,a_m,s'_m \rangle \right \}_{m=1}^{M} $ and a query set $\mathcal{T}_{e_i}^{qry} = \left \{ \langle \bar{s}_n,\bar{a}_n,\bar{s}'_n\rangle \right \}_{n=1}^{N} $. 
Afterward, the local dynamics model $\mathcal{D}_{\theta_i = \theta_{\mathcal{S}}}$ is trained on the support set $\mathcal{T}_{e_i}^{spt}$ using a gradient-descent learning algorithm, resulting in updated parameters $\theta_i^{'}$.
Then, $\mathcal{D}_{\theta_i^{'}}$ is evaluated on the query set $\mathcal{T}_{e_i}^{qry}$, and a loss value $\mathcal{L}(\mathcal{T}_{e_i}^{qry}, \mathcal{D}_{\theta_i^{'}})$ is calculated as a consequence.
% Note that the local loss reflects the learning ability of $\mathcal{D}_{\theta_i}$ which is parameterized by $\theta_{\mathcal{S}}$. 
Note that this loss value can reflect the capability of identifying the hyper-parameters, i.e., the learning ability of $\mathcal{D}_{\theta_i = \theta_{\mathcal{S}}}$.
%%
% step-3 & step-4
At the third and the forth step, 
%the server aggregates all the loss values from federated RL systems, computes the gradient and then updates the server dynamics model $\mathcal{S}$.
% Mathematically, 
loss values $\left \{ \mathcal{L}(\cdot, \theta_i^{'}) \right \}_{i=1}^{K} $ are collected by the server from all the federated RL systems, which are utilized for optimizing the server dynamics model $\mathcal{S}(\theta_\mathcal{S})$ as the following objective:
% Based on these loss values, the server dynamics model $\mathcal{S}(\theta_\mathcal{S})$ are optimized to minimize the aggregated loss as the following objective:
\begin{equation}
        \min_{\theta_{\mathcal{S}}} \frac{1}{K}  \sum_{i=1}^{K} \mathcal{L}(\mathcal{T}^{qry}_{e_i}, \mathcal{D}_{\theta_i^{'}})  \quad \text{s.t.} \quad
        \theta_{i}^{'} = \theta_{i} - \alpha \nabla_{\theta_i} \mathcal{L} (\mathcal{T}^{spt}_{e_i}, \mathcal{D}_{\theta_{i}=\theta_{\mathcal{S}}}),
    \label{eq:preparation-objective}
\end{equation}
where $\alpha$ is the learning rate. 
Here, the loss function $\mathcal{L}$ can be defined as Mean Square Error (MSE) in continuous state domains or as Cross Entropy in discrete state domains.
When the parameters $\theta_\mathcal{S}$ are optimized, 
the server dynamics model $\mathcal{S}$ is a function parameterized by $\theta_\mathcal{S}^{*}$. 
% We consider $\mathcal{S}(\theta_\mathcal{S}^{*})$ as being representative of the knowledge of the training-environment structure, which is capable of quickly personalize to an environment with specific hyper-parameters. 
$\mathcal{S}(\theta_\mathcal{S}^{*})$ is considered representative knowledge of the training-environment structure, capable of quickly personalizing to an environment with specific hyper-parameters.

\paragraph{Diagnosis:}

%% diagnosis
Diagnosis is a time-sensitive and resource-sensitive stage, which is carried out prior to the deployment of the learned policy.
Namely, a poisoned agent is expected to quickly understand the dynamics of its deployment environment based on a small number of interactions. 
To achieve this, the agent is provided with the critical knowledge of environment structure and accordingly identify the environment dynamics. 
% Specifically, the agent retrieves knowledge from the server and personalizes it based on its deployment environment.
Such a diagnosis stage consists of three steps:
(1) the retrieval of knowledge (i.e., parameters) from the server dynamics model;
(2) the collection of trajectories in the deployment environment;
(3) the understanding of the deployment environment (i.e., dynamics model).

%% 数学描述
Imagine an RL agent which has learned its policies in a poisoned training environment and aims to accurately grasp the dynamics of the deployment environment parameterized by $\hat{e} \sim p(e)$.
The agent first initializes its local dynamics model using the parameters $\theta_{\mathcal{S}}^{*}$ retrieved from the server dynamics model $\mathcal{S}$.
Then, the local dynamics model $\mathcal{D}_{\theta_{\hat{e}} = \theta_{\mathcal{S}}^{*}}$ is fine-tuned based on a small set of transitions $\mathcal{T}_{\hat{e}} = \left\{ \langle s_n, a_n, s_{n+1}  \rangle \right\}_{n=1}^{N}$ sampled in the deployment environment, denoted as
% $\theta_{\hat{e}}' = \theta_{\hat{e}} - \alpha \nabla_{\theta_{\hat{e}}} \mathcal{L} (\mathcal{T}_{\hat{e}}, \mathcal{D}_{\theta_{\hat{e}}=\theta_{\mathcal{S}}^{*}}).$
\begin{equation}
    \theta_{\hat{e}}' = \theta_{\hat{e}} - \alpha \nabla_{\theta_{\hat{e}}} \mathcal{L} (\mathcal{T}_{\hat{e}}, \mathcal{D}_{\theta_{\hat{e}}=\theta_{\mathcal{S}}^{*}}).
\end{equation}
Note that the diagnosis performance relies on the knowledge quality retrieved from the server, i.e., the learning ability of the server dynamics $\mathcal{S}$ parameterized by $\theta_{\mathcal{S}}^{*}$.

\paragraph{Recovery:}
%% recovery
Recovery is the final stage of the policy-resilience mechanism, which aims to reduce the negative effects of the EPA and restore policy deployment performance to the full extent of its ability.
The policy recovery is accomplished using the imagined trajectories derived from the dynamics model $\mathcal{D}_{\theta_{\hat{e}}^{'}}$ that has been tailored to the deployment environment.
% during the diagnosis stage.
% Specifically, the imagined trajectories are generated by the dynamics model $\mathcal{D}_{\theta_{\hat{e}}^{'}}$ that has been tailored to the deployment environment during the diagnosis phase.
The success of recovery are generally applicable for both model-free and model-based RL agent. 
For example, a model-free (MF) RL agent can recover its policy by adopting model-based value expansion \cite{feinberg2018model, buckman2018sample} while a model-based (MB) RL agent may choose Model Predictive Control with cross-entropy method  \cite{de2005tutorial} for restoring its policy. 
% Details refer to Appendix \ref{appendix-methodology}.

%%%%%%%%%%%%%%%%%%%%%%%%%%%%%%%%%%%%%%%%%%%%%%%%%%%%%%
\section{Experiments}  %%%%%%%%%%%%%%%%%%%%%%%%%%%%%%%%
%%%%%%%%%%%%%%%%%%%%%%%%%%%%%%%%%%%%%%%%%%%%%%%%%%%%%%

% \begin{wrapfigure}{r}{0.5\textwidth}
%     \begin{minipage}{0.5\textwidth}
%         \vspace{-0.8cm}
%             \begin{figure}[H]
%                 \centering
%                 \includegraphics[width=0.8\linewidth]{Figures/3D_mitigation_trend.pdf}
%                 \caption{Evaluation of policy resilience.}
%                 \label{fig:mitigation_trend}
%             \end{figure}
%         \vspace{-0.8cm}
%     \end{minipage}
% \end{wrapfigure}

\begin{figure}[ht]
    \centering
	\begin{minipage}[ht]{0.32\linewidth}
	    \vspace{0.55cm}
        \includegraphics[width=\linewidth]{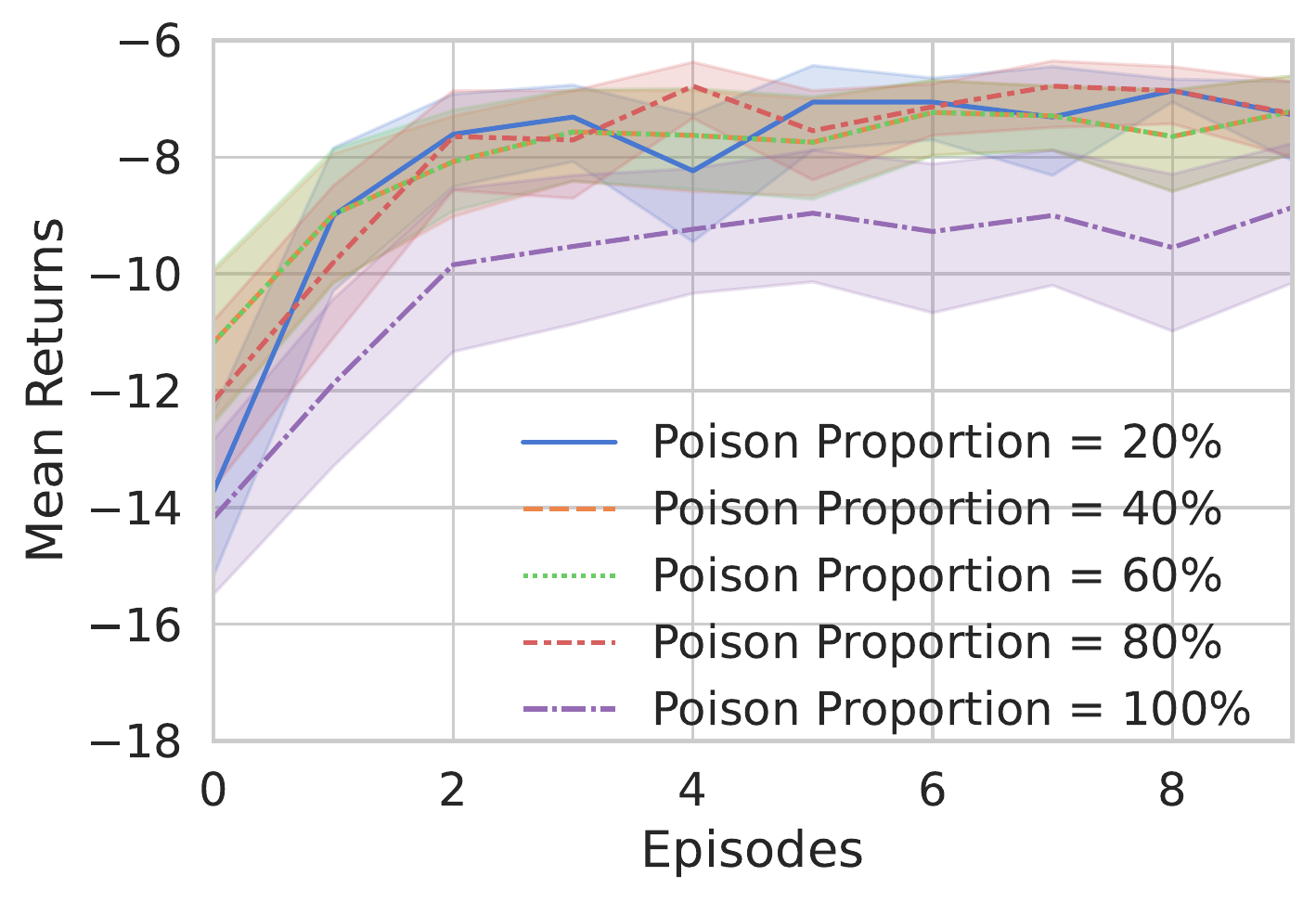}
        \caption{3D Grid world}
        \label{fig:3D_cmp}
        \vspace{-0.2cm}
	\end{minipage}
	\begin{minipage}[ht]{0.32\linewidth}
        \includegraphics[width=\linewidth]{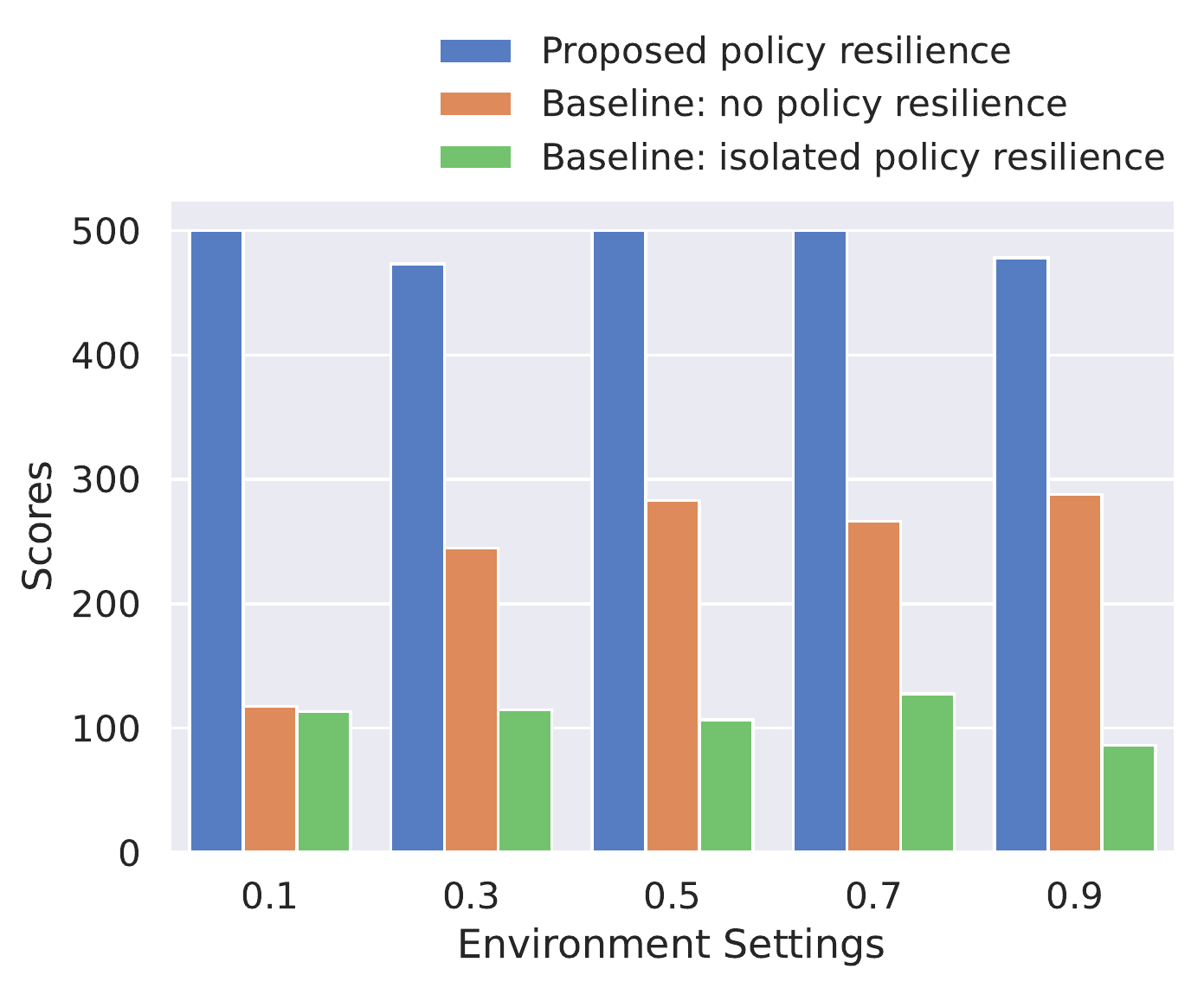}
        \caption{Cartpole, MF}
        \label{fig:cartpole_mve}
        \vspace{-0.2cm}
	\end{minipage}
	\begin{minipage}[ht]{0.32\linewidth}
        \includegraphics[width=\linewidth]{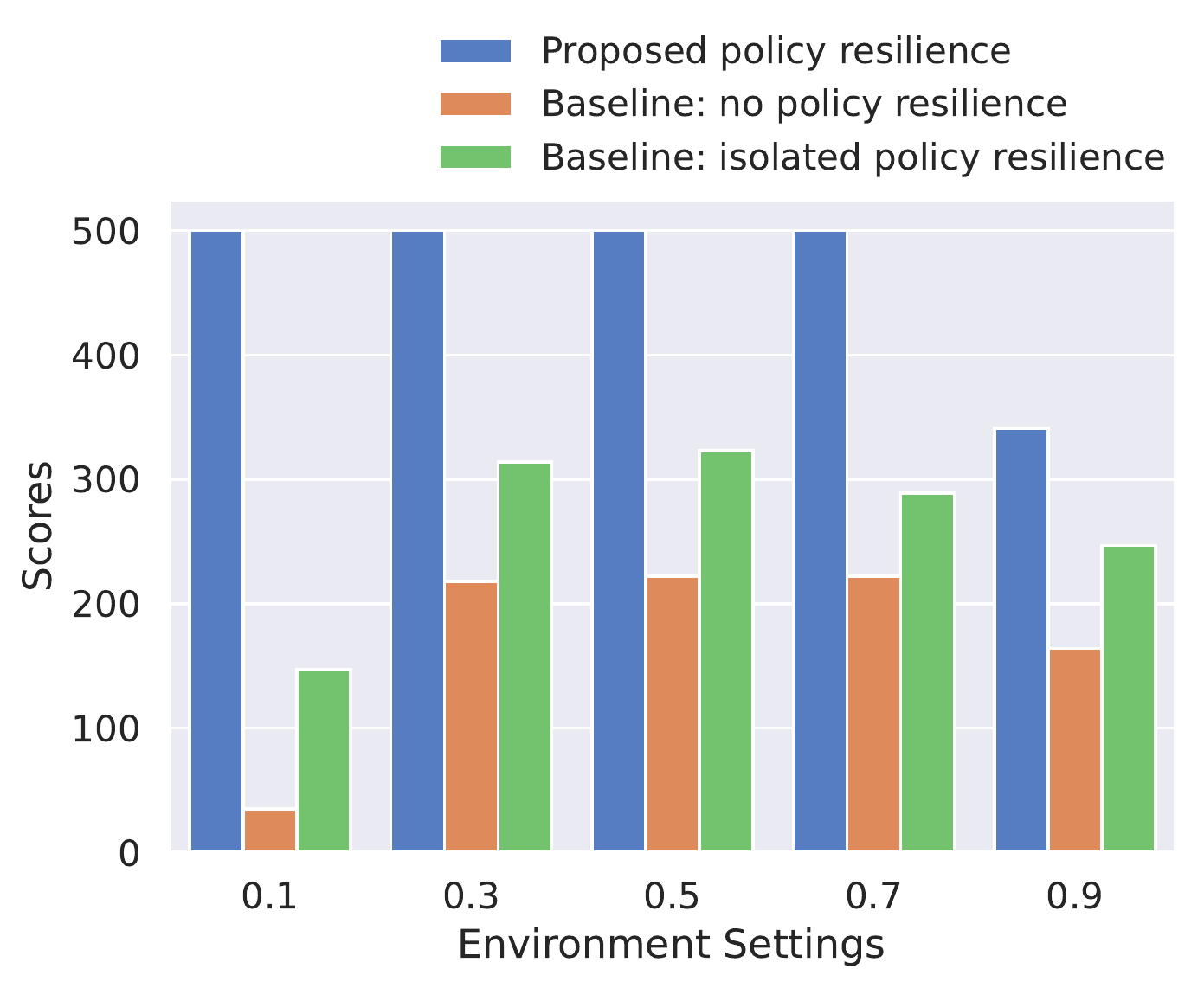}
        \caption{Cartpole, MB}
        \label{fig:cartpole_mpc}
        \vspace{-0.2cm}
    \end{minipage}
\end{figure}

We evaluate the recovery of a poisoned RL policy in a 3D grid world, when different proportions of poisoned agents are present at the preparation stage. 
Figure \ref{fig:3D_cmp} shows that poisoned policies are effectively recovered to good performance within only two-episode interactions. 
This means that the policy recovery is time- and resource-efficient due to the knowledge sharing.
We further evaluate the applicability of our policy-resilience mechanism, showing the successful policy recovery on agents adopting either model-free (MF) or model-based (MB) RL learning algorithms, as shown in Figure \ref{fig:cartpole_mve} and \ref{fig:cartpole_mpc}. 
More results and discussions refer to Appendix \ref{appendix-experiment}.

% It means that the server dynamics model, which is learned using meta-learning within the design of federated Hip-MDP systems, represents critical knowledge of the common underlying environment structure. 
% % extracting environment information from a set of client RL agents, provides beneficial knowledge to the agent. 
% Such knowledge facilitates the agent to quickly identify the dynamics of the deployment environment, which allows the policy recovery using imagined trajectories.
% % More evaluations and discussions can be found in Appendix.

%%%%%%%%%%%%%%%%%%%%%%%%%%%%%%%%%%%%%%%%%%%%%%%
\section{Conclusion}   %%%%%%%%%%%%
%%%%%%%%%%%%%%%%%%%%%%%%%%%%%%%%%%%%%%%%%%%%%%%

The aim of this paper is to investigate resilience of RL policies against the training-time environment poisoning attacks.
A policy-resilience mechanism is proposed to recover a poisoned policy for an optimal deployment performance, which is described as preparation, diagnosis and recovery.
It is designed as a federated framework incorporated with a meta-learning approach, allowing efficient extraction and sharing of environment knowledge.
When armed with shared knowledge, a poisoned agent can effectively identify the dynamics model of the deployment environment from limited interactions.
In this way, imagined trajectories can be derived from the dynamics model, which is then used to recover the deployment performance of the poisoned policy.
% We summarize such a policy resilience as three stages, namely preparation, diagnosis, and recovery.
% Based on the shared knowledge, the federated RL agent can efficiently understand the deployment environments and thereby restore the corrupted policy.
% Such a policy-resilience mechanism is described as a three-step procedure, namely preparation, diagnosis and recovery.
% %
Our empirical evaluation shows the effectiveness and efficiency of our policy resilience to EPA on both model-free and model-based RL agents.

\bibliographystyle{ieeetran.bst}
\bibliography{ref.bib}

\newpage
\appendix
% \section{Appendix}

% %% ----- figures: examples 
% \begin{figure*}[ht]
%     \centering
%     \includegraphics[width=\linewidth]{Figures/pipeline_illustration.pdf}
%     \caption{An example of policy-resilience mechanism in RL systems.}
%     \label{fig:FRL_examples}
% \end{figure*}

% %% --- figure: robustness and resilience ---
% \begin{figure*}[ht]
%     \centering
%     \includegraphics[width=0.6\linewidth]{Figures/Resilience_vs_Robust.pdf}
%     \caption{Illustration of robustness and resilience.}
%     \label{fig:cmp_robust_resilience}
% \end{figure*}

%%% ===============================================================
\section{Related Works}
% 背景 + 动机（研究 resilience 的动机）
To guarantee the security associated with the learning of RL policies, defences against training-time attacks have been developed from the standpoint of {\em robustness} which refers to the ability of an agent to maintain its functionality in the presence of perturbations \cite{behzadan2017whatever}.
There are two general categories of these works:
(1) Some works \cite{banihashem2021defense, lykouris2021corruption, chen2021improved, wei2022model, zhang2021robustrl, zhang2021robust, wu2022copa} offer theoretical guarantees concerning the policy learning under perturbations within a certain range;
(2) Other studies \cite{behzadan2017whatever, behzadan2018mitigation, wang2020reinforcement, zhang2021robust} empirically examine the robustness of the training model to perturbations.
In spite of the fact of robustness is a crucial issue, it is merely an add-on concern when designing RL algorithms, which could increase design costs or compromise policy performance.  
Thus, there may be a lack of resources or priority in incorporating robustness into the design of RL systems. 
In addition, a successful robustness must be prepared to defend against all vulnerabilities and their associated attacks.
However, the majority of vulnerabilities are unknown until they are exploited by an attacker, or known but unable to be prevented in advance.
As a result, it is difficult and costly to achieve complete robustness, especially when attack approaches have been developed intensively.
In light of these constraints, relying solely on robustness mechanisms is insufficient to protect the learning of RL policies from being poisoned.
There is a need to shift the focus of security from robustness to {\em resilience} which in this context refers to an RL agent's ability to recover its policy from malicious manipulations \cite{behzadan2017whatever}.
However, very few studies focus on the defenses against training-time attacks from a resilience perspective. 
In this work, we attempt to study a policy-resilience mechanism against environment-poisoning attacks on RL policy learning.

%%% ===============================================================
\section{Preliminaries}
\label{appendix-preliminaries}

\paragraph{Hidden-Parameter Markov Decision Process.}

A Hidden Parameter MDP (HiP-MDP) \cite{doshi2016hidden} represents a family of MDPs in which hidden parameters $e \in \mathbb{R}^n$ are used to parameterize the transition dynamics.
As examples of hidden parameters, we can mention gravity, friction on a surface or the strength of a robot actuator. These parameters are not part of the observation space but play a significant role in the response of the environment to the actions of agents \cite{perez2020generalized}. 
Note that the hidden parameter shares the same definition of hyper-parameters in TEPA and DBB-EPA, thereby we use the expression hyper-parameter uniformly in the following.
Formally, a HiP-MDP can be defined as a tuple $\langle S,A,R, F_e, \gamma \rangle$, in which $S$ is the set of states, $A$ is the set of actions, and $R$ is a reward function.
$F(s'|s,a,e_i)$ is a transition function for each task instance $i$ parameterized by the hyper-parameter $e_i$. Here, the parameter $e_i$ is drawn from a prior distribution $e_i \sim p(e)$. 
The Hip-MDP framework assumes that variations in the dynamic of true tasks can be fully described by a finite-dimensional array of hidden parameters \cite{doshi2016hidden}. 

\paragraph{Federated Learning.}

%% what is federated learning
The main idea of Federated Learning (FL) \cite{mcmahan2017communication} is to utilize distributed data sets to generate machine learning models with protection of data privacy and security. 
FL succeeds to solve the dilemma that the data is distributed at isolated islands but is forbid to be collected/fused for processing model \cite{yang2019federated}.
% FRL application 
Recently, FL has addressed data isolation and privacy issues in RL-based applications, such as smart energy management \cite{lee2020federated}, autonomous driving in Internet of Vehicles (IoV) \citep{liang2019federated} and optimal control of internet-of-things devices (IoT) \citep{lim2020federated}.

\paragraph{Meta Learning.}

%% what is meta-learning
The meta-learning method involves learning a model from an array of tasks in order to make it capable of solving new tasks with only a limited number of training examples \cite{finn17maml}.
Using meta-learning, a parameterized algorithm (i.e., meta-learner) is learned by performing a meta-training process, and it can be used to fast train a specific model in each task.
%% how to update parameterized algorithms
Specifically, each task consists of two distinct data sets: a support data set and a query data set.
A task-specific model is trained on the support set and then tested on the query set. Using these test results, the parameterized algorithm is updated.
It is important to note that the parameterized algorithm is capable of learning how to adapt to new tasks more quickly, i.e., "learning to fine-tune".

% %% --- figure: pipeline ---
% \begin{figure*}[ht]
%     \centering
%     \includegraphics[width=0.8\linewidth]{Figures/resilience_pipeline.pdf}
%     \caption{Pipeline of policy-resilience mechanism.}
%     \label{fig:pipeline}
% \end{figure*}

%%% =================================================================================

\section{Methodology}
\label{appendix-methodology}

% %% -------------------------------------------
% \subsection{Examples of Federated RL System}

% In our policy-resilience mechanism, isolated RL systems are organized in a federated manner, which are common in real-world applications. 
% %
% For instance, autonomous vehicles could be organized by a transportation agency in a federated manner.
% As a center server, the transportation agency can collect environment data (e.g., road conditions) from a variety of vehicles without compromising data privacy, and extract useful environment knowledge to share with each vehicle to assist it driving in different environment instances. 
% Similar examples include robots connected via a cloud server, buildings organized by a smart city center, and financial companies supervised by a central bank. 

%% -----------------------------------------
\subsection{Details of Preparation Stage}

\begin{figure*}[ht]
    \centering
    \includegraphics[width=\linewidth]{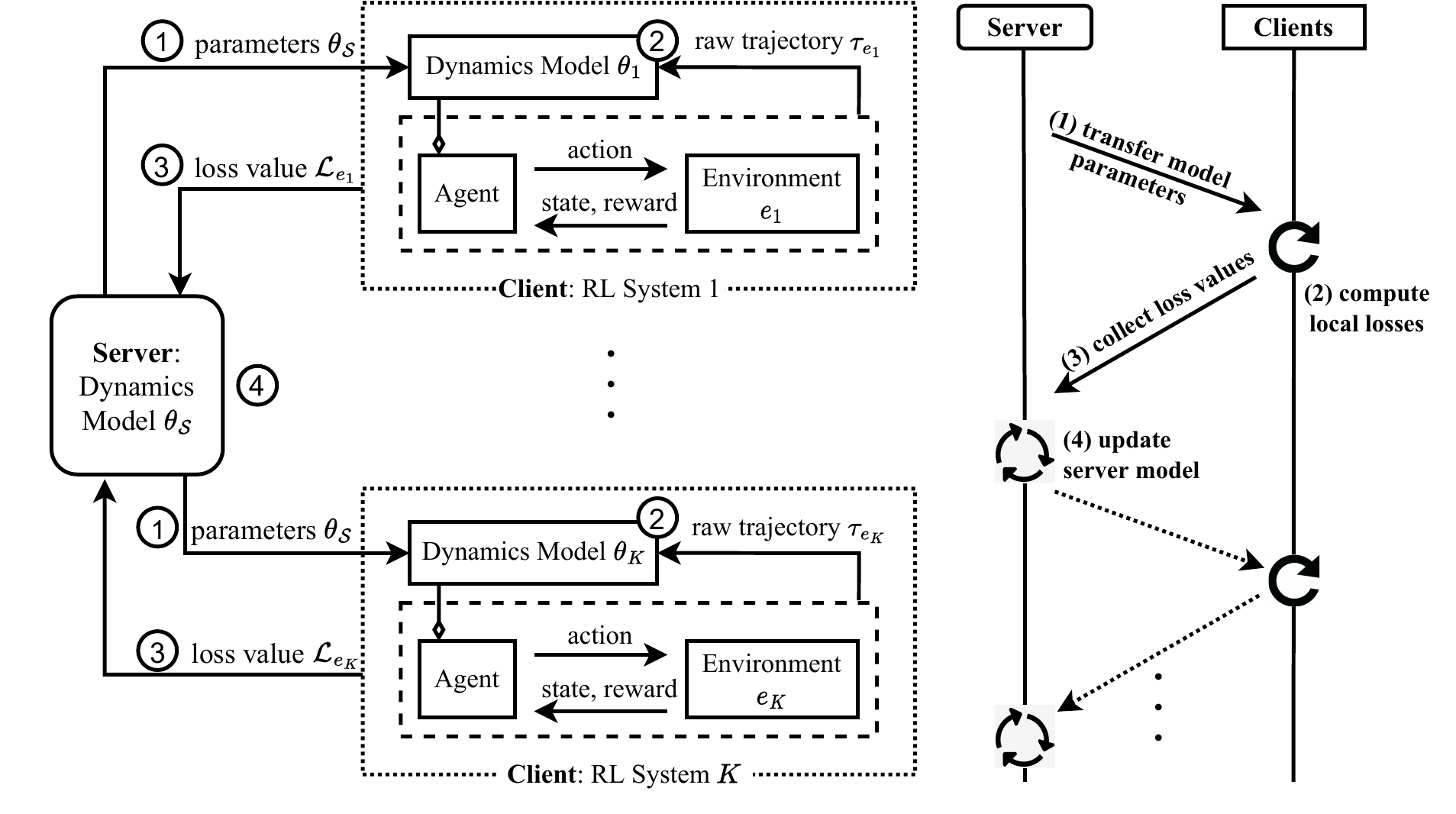}
    \caption{Illustration of preparation stage in policy-resilience mechanism.}
    \label{fig:resilience_framework}
\end{figure*}

Figure \ref{fig:resilience_framework} illustrates the learning of the server dynamics function based on environment information collected from all the client RL systems. 

%% --- figure: preparation & framework ---

The server dynamics model is learned through a combination of federated learning and meta-learning approaches \cite{finn17maml}, which is finalized as Algorithm \ref{Algo:fusion}. 

% ------------ Algorithm : double-black-box ---------------
\begin{algorithm}[ht]
\centering
\caption{Preparation: Learning of the Server Dynamics Model $\mathcal{S}$}
\begin{minipage}{\textwidth}
\begin{algorithmic}[1]
    % \Require A server dynamics model $\mathcal{S}$
    \Require Client RL systems $\{\mathcal{C}_i\}_{i=1}^{K}$ 
    \State Initialize the global dynamics model $\mathcal{S}(\theta_{\mathcal{S}})$
    \For{{n\_round} = 1,2,...}
        % modeling
        \For{each client $\mathcal{C}_i$ in parallel}
            \State Initialize $\mathcal{D}_i$ with $\theta_{\mathcal{S}}$ retrieved from $\mathcal{S}$
            % sampling
            \If {{ n\_round} mod { n\_interval} = 0}
                \State Collect $X$ transitions $\langle s,a,s' \rangle$ 
                \State $\mathcal{T}_{e_i} \leftarrow \mathcal{T}_{e_i} \cup \{ \left \langle s_j,a_j,s_j'  \right \rangle\}_{j=1}^{X}$
            \EndIf
            \State Sample $\left \{\mathcal{T}^{spt}_{e_i}, \mathcal{T}^{qry}_{e_i} \right \}$ from $\mathcal{T}_{e_i}$
            \State Update local parameters $\theta^{'}_i \leftarrow \theta_{i} - \alpha \nabla_{\theta_i} \mathcal{L} (\mathcal{T}^{spt}_{e_i}, \mathcal{D}_{\theta_{i}=\theta_{\mathcal{S}}})$
            \State Compute loss values $\mathcal{L}(\mathcal{T}^{qry}_{e_i}, \mathcal{D}_{\theta_i^{'}})$ following Eq.[\ref{eq:loss_continuous}] or Eq.[\ref{eq:loss_discrete}]
            \State Send the loss value $\mathcal{L}(\cdot, \theta^{'}_i)$ to server $\mathcal{S}$
        \EndFor
    % server update
    \State In server, aggregate losses $\mathcal{L}_{\mathcal{S}} \leftarrow \frac{1}{K} \sum_{i=i}^K \mathcal{L}(\cdot, \theta^{'}_i)$
    \State \hspace{0.7cm} update parameters $\theta_{\mathcal{S}}^{'} \leftarrow \theta_{\mathcal{S}} - \beta \nabla_{\theta_{\mathcal{S}}} \mathcal{L}_{\mathcal{S}}$
    \EndFor
    \State Return parameters $\theta_{\mathcal{S}}^{*}$ of global dynamics model $\mathcal{S}$
\end{algorithmic}
\end{minipage}
\label{Algo:fusion}
\end{algorithm}

% definition of loss function
Additionally, we provide a specific definition of the loss function $\mathcal{L}(\mathcal{T}_e, \mathcal{D}_{\theta})$ depending on the type of environment state domain, i.e. continuous state domains and discrete state domains.
For example, in continuous state domains, we measure the loss value via Mean Square Error (MSE) as
\begin{equation}
    \mathcal{L} (\mathcal{T}_{e}, \mathcal{D}_{\theta}) = \sum_{j=0}^{X} \left\| \mathcal{D}_{\theta} ({s}_j, {a}_j) - {s}_{j+1} \right\|^{2}_{2},
    \label{eq:loss_continuous}
\end{equation}
where $X$ is the number of transitions in $\mathcal{T}_{e}$.
Similarly, the loss value is measured by cross entropy in discrete state domains, which is denoted as
\begin{equation}
    \mathcal{L} (\mathcal{T}_{e}, \mathcal{D}_{\theta}) = - \mathbb{E}_{\langle s, a, s' \rangle \sim \mathcal{T}_{e}} \log \mathcal{D}_{\theta}(s'|s,a)
    \label{eq:loss_discrete}
\end{equation}

%% -----------------------------------------
\section{Details in Recovery Stage}
%% 目的
The goal of the recovery stage is to restore the deployment performance of poisoned policies to their full potential.
The recovery of policy depends on imagined trajectory, therefore an understanding of the dynamics of the deployment environment (i.e., diagnosis) is crucial to finding an optimal policy.
We present possible recovery solutions from the perspective of a model-free RL agent and a model-based RL agent, respectively.

%% dynamics model 的作用
To begin with, we describe the imagined trajectories generated by the dynamics model $\mathcal{D}_{\theta_{\hat{e}}^{'}}$ that has been tailored to the deployment environment during the diagnosis phase. 
In a deterministic environment, an agent utilizes the dynamics model to predict the future state $s'$ based on the current state $s$ and the chosen action $a$, denoted as $s' \sim \mathcal{D}_{\theta_{\hat{e}}^{'}} (s,a)$.
% \begin{equation*}
%     s' \sim \mathcal{D}_{\theta_{\hat{e}}^{'}} (s,a).
% \end{equation*}
%
In a stochastic environment, the dynamics model may be designed as an uncertainty-aware neural network  \cite{gal2016dropout, lakshminarayanan2017simple}, which consists of ensembles of independently trained models rather than a single model.
The agent can predict the future states with consideration of uncertainty which is estimate via the mean and deviation of multiple stochastic forward passes of the models. The state prediction is denoted as $s^{'} \sim \mathcal{N} \left(\mathbb{E} \left[\mathcal{D}_{\theta_{\hat{e}}^{'}}(s, a) \right], \sqrt{\text{Var} \left[ \mathcal{D}_{\theta_{\hat{e}}^{'}}(s, a) \right]} \right)$ where $\mathcal{N}$ represents Gaussian distribution.
Therefore, in both deterministic and stochastic environment, the agent is able to recursively predict future states $\left\{ s_{t+1}, \dots, s_{t+H} \right \}$ via a action sequence $\{a_t, \dots, a_{t+H-1} \}$ given an initial state $s_0$, generating imagined trajectories. 

%% model-free agent
\paragraph{Model-Free RL Agent.}
For a model-free RL agent that attempts recover its policy using imagined trajectory, we adopt a model-based value expansion (MVE) \cite{feinberg2018model, buckman2018sample} as the approach 
% to incorporating the learned dynamics model $\mathcal{D}_{\theta_{\hat{e}}^{*}}$ into its model-free value function estimation. 
The MVE approach incorporates the dynamics model $\mathcal{D}_{\theta_{\hat{e}}^{'}}$ into value estimation by replacing the standard Q-learning target with an improved one $V_h^{MVE}$. 
$V_h^{MVE}$ is computed by rolling the dynamics model $\mathcal{D}_{\theta_{\hat{e}}^{'}}$ out for $h$ steps, which limits imagination to a fixed depth $h$ so as to prevent accumulative errors due to inaccuracies in the dynamics model. 
Specifically, the value estimation $V_h^{MVE}(r,s')$ is composed of a short-term value estimate derived from the unrolling of the dynamics model $\mathcal{D}_{\theta_{\hat{e}}^{'}}$, and a long-term value estimate derived from the learned values $Q^{\pi}_{\theta^{-}}$, denoted as
\begin{equation}
    V_h^{MVE}(r,s') = r + \left( \sum^{h}_{i=1} T^{i} \gamma^{i} R(s_{i-1}', a_{i-1}', s_{i}') \right) + T^{h+1} \gamma^{h+1} Q^{\pi}_{\theta^{-}}(s_h', a_h'),
    \label{eq:recover_mf}
\end{equation}
where $s_i \sim \mathcal{D}_{\theta_{\hat{e}}^{'}} (s_{i-1}, a_{i-1})$ and $Q^{\pi}_{\theta^{-}}$ is an approximated action-value function. 
$T$ is the termination of the trajectory and $R$ represents a reward function, which are assumed to be known in this context. Note that $T$ and $R$ also can be learned based on trajectories.

%% model-based agent
\paragraph{Model-Based RL Agent.}
The learning algorithm adopted by a model-based RL agent is Model Predictive Control (MPC) with cross-entropy method (CEM) \cite{de2005tutorial}. 
MPC is an online learning approach which can re-plan an action based on updated state information, so that it can prevent accumulative errors caused by the dynamics model $\mathcal{D}_{\theta_{\hat{e}}^{'}}$. 
Specifically, at time step $t$, the MPC approach iteratively samples action $a_{t:t+H-1}$ from multivariate normal distributions $\mathcal{N}(a_t|\mu_t, \sigma_t)$ which is adjusted based on the best sampled actions. Here, $\mu_t$ and $\sigma_t$ are the mean and the variance of top 10\% actions. 
In MPC, the agent implements the first action from the optimal action sequence $a_{t:t+H-1}$ and then optimally re-plans an action sequence at time step $t+1$.
Finally, the agent aims to find an action sequence which optimizes 
\begin{equation}
    \sum^{t+H-1}_{t} \mathbb{E}_{s_{t+1} \sim \mathcal{D}_{\theta_{\hat{e}}^{*}} (s_t, a_t)} \big[ r(s_t, a_t, s_{t+1}) \big]
    \label{eq:recover_mb}
\end{equation}
using predicted state $s_{t+1}$ from the local dynamics model $\mathcal{D}_{\theta_{\hat{e}}^{*}}$.

Based on the solutions outlined above, either a model-free or a model-based agent can recover its poisoned policy using imagined trajectories.
Algorithm \ref{Algo:policy_correction} summarizes the operations performed during the diagnosis and recovery stages.

%% ------ recover of poisoned policy ------ %%
\begin{algorithm}[ht]
\centering
\caption{Diagnosis and Recovery}
\begin{minipage}{\textwidth}
\begin{algorithmic}[1]
    \Require An optimized server dynamics model $\mathcal{S}(\theta_{\mathcal{S}}^{*})$
    \Require An RL system with dynamics model $\mathcal{D}_{\theta_{\hat{e}}}$ and poisoned policy $\pi_{\hat{e}}$
    \State Initialize $\mathcal{D}_{\theta_{\hat{e}}}$ with $\theta_{\mathcal{S}}^{*}$ retrieved from $\mathcal{S}$  
    \State Collect $n$-episode transitions into data set $\mathcal{T}_{\hat{e}}$
    \State Update parameters of $\mathcal{D}_{\theta_{\hat{e}}}$ as $\theta_{\hat{e}}' = \theta_{\hat{e}} - \alpha \nabla_{\theta_{\hat{e}}} \mathcal{L} (\mathcal{T}_{\hat{e}}, \mathcal{D}_{\theta_{\hat{e}}=\theta_{\mathcal{S}}^{*}})$
    \State Recover $\pi_{\hat{e}}$ following Eq. \ref{eq:recover_mf} or \ref{eq:recover_mb}, using trajectories imagined via $\mathcal{D}_{\theta_{\hat{e}}^{'}}$
\end{algorithmic}
\end{minipage}
\label{Algo:policy_correction}
\end{algorithm}

%% =============================================================================
\section{Experiments}
\label{appendix-experiment}

We empirically evaluate our proposed policy-resilience mechanism by verifying the following questions: 
(1) Can our approach recover the poisoned RL policy from malicious manipulation caused by corrupted training environments?
(2) Does the sharing of environment knowledge enable a poisoned agent to grasp the dynamics of deployment environments more efficiently for policy recovery?
(3) Does the meta-learning mechanism provide an effective means of extracting critical knowledge about environment dynamics within a federated framework?
In this section, we provide empirical answers to these questions in both discrete and continuous state domains.  

% Does the meta-learning mechanism protect the merged knowledge in the server from being negatively impacted by client RL systems in which the training environment has been contaminated?

%% ----------------------------------------------------- %%
\subsection{Discrete State Domains} %% ----------------- %%
%% ----------------------------------------------------- %%

This part evaluates the design and performance of a policy-resilience mechanism in discrete state domains.

\subsubsection{Experiment Settings}

Our proposed policy-resilience mechanism is targeting the environment-poisoning attacks at training time, particularly those attacks which poison the environment dynamics by perturbing the parameters of the physical environment (i.e., the hyper-parameters). 

%% 详细描述 攻击 的方式
\paragraph{Environment:}

The environment-poisoning attack is implemented on an RL agent performing navigation tasks in one $5 \times 5$ grid world as shown in Figure \ref{fig:3D_grid}. 
The experiment environment (i.e., 3D grid world) is the same as that of TEPA, which simulates mountains or rugged terrain.
In this 3D grid world, the agent's success of moving from one cell to the neighboring one is proportional to their relative elevation, changes in elevation will influence how the environment responds to the agent's action (i.e., environment dynamics). 
Therefore, the elevation setting is regarded as the hyper-parameters which can be tweaked by the attacker.

\paragraph{Baselines:}
Due to the limited number of studies that examine policy resilience against environmental poisoning, there is no appropriate baseline that can be directly referenced.
Thus, we design two types of baselines in accordance with our experimental objectives. 
\begin{itemize}
    \item Policy-Resilience Mechanism without Federated Framework: 
    To evaluate whether the federated structure is necessary for extracting and sharing environment knowledge, we design a baseline policy-resilience mechanism which does not rely on server-based environment knowledge. 
    Thus, both during the preparation and diagnosis stage, the agent learns the environment dynamics model locally without shared knowledge, using local Stochastic Gradient Descent (SGD) updates.
    This baseline is termed as {\em No Federated Framework}.
    \item Policy-Resilience Mechanism without Meta-learning Mechanism: 
    To evaluate the effect of a meta-learning mechanism on the quality of environment-knowledge extraction and fine-tuning, we design a baseline policy-resilience procedure where the federal dynamics model is learned using Federated Averaging Learning \cite{mcmahan2017communication}. 
    It means that during the preparation process, the federal dynamics model is optimized by averaging local SGD updates.
    This baseline is termed as {\em No Meta-Learning Mechanism}.
\end{itemize}

%% ============================================================================

\subsubsection{Experiment Results}

For these experiments, the policy-resilience framework includes 10 client RL systems that are all assumed to belong to the same Hip-MDP.
We initialize the hyper-parameters of these RL environments at the same values in order to simplify the discussion.
All of them are involved in the preparation process, and some of them may be poisoned during training.

\begin{wrapfigure}{r}{0.5\textwidth}
    \begin{minipage}{0.5\textwidth}
        \vspace{-0.8cm}
            \begin{figure}[H]
                \centering
                \includegraphics[width=0.8\linewidth]{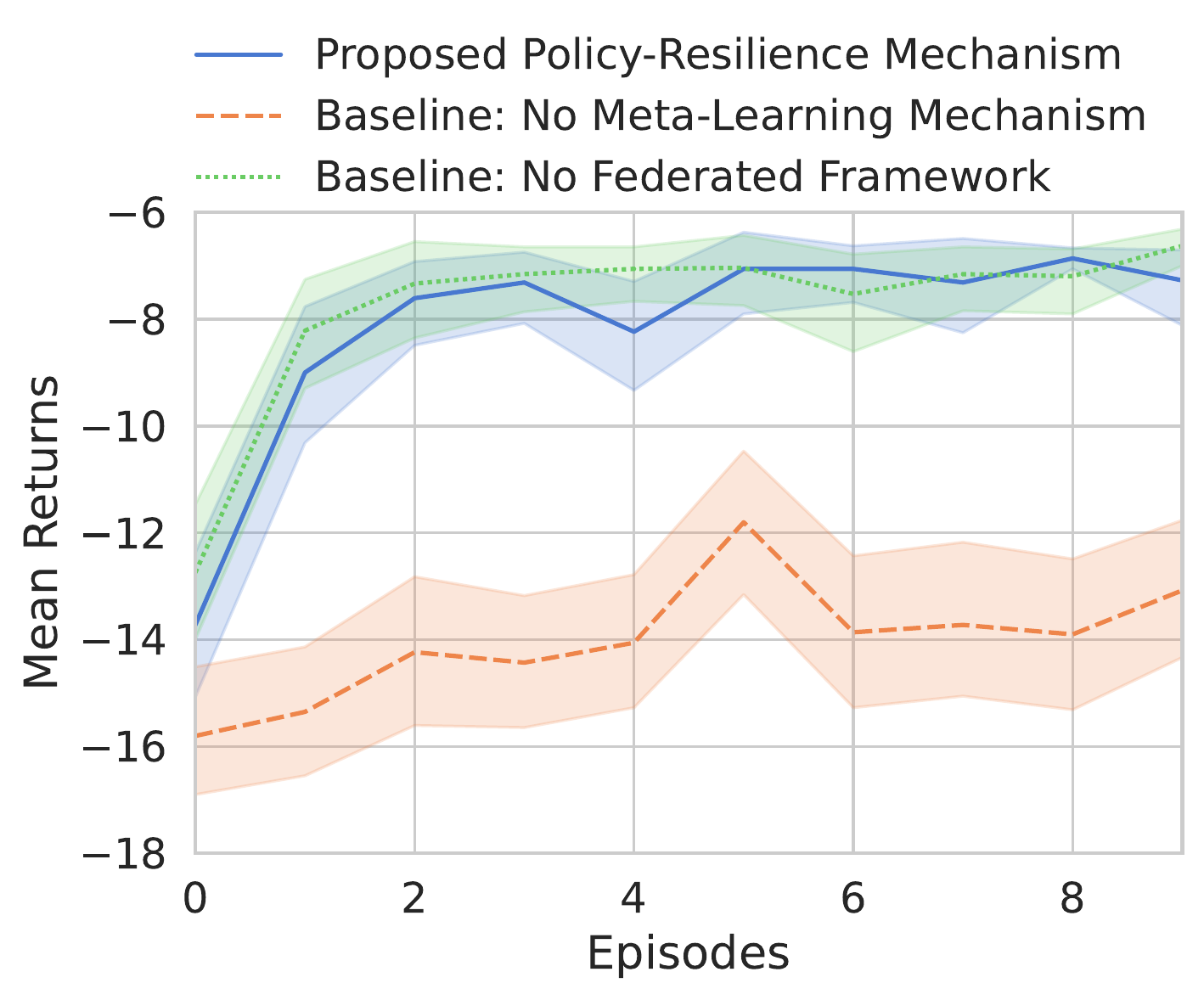}
                \caption{Performance comparison.}
                \label{fig:3D_grid}
            \end{figure}
        \vspace{-0.3cm}
    \end{minipage}
\end{wrapfigure}

\paragraph{Effect of Federated Design.} %% ----------------------
% 实验目的 或 背景
We are investigating the effects of the shared knowledge which is achieved by the design of a federated framework.
In this experiment, only one RL system has been attacked at the preparation stage and it transfers the information of the corrupted environment to the server.
% 图片-3 描述 或 说明
Figure \ref{fig:3D_grid} shows the performance comparison between our policy-resilience mechanism and the baseline {\em No Federated Framework}). 
As shown, our approach succeeds in recovering the performance of a poisoned policy based on a dynamics model that is initialized by the shared knowledge (i.e. parameters of server dynamics model) and fine-tuned using the local two-episode trajectory data.
In contrast, the baseline {\em No Federated Framework}) fails in recovering the policy when the agent is trying to improve its local dynamics model without the shared environment knowledge.  
% 实验结论
These results demonstrate that shared knowledge of the deployment environment allows the agent to quickly and accurately understand the deployment environment, enabling imagined trajectories to be generated for policy recovery.
The knowledge extraction and sharing require federated frameworks, so federated frameworks are essential and effective design elements.

Actually, we have designed our policy-resilience mechanism in such a way that an agent in our RL will not need to invest additional resources or add-on capabilities for achieving resilience.
Any agent may be provided with information on the environment and allowed to recover its poisoned policies accordingly, regardless of whether it participated in the preparation process.
Thus, in the context of a single RL system, such a policy-resilience mechanism is resource-efficient.

\paragraph{Effect of Meta-Learning Design.} %% ----------------------
% 实验目的 或 背景
In this experiment, we aim to evaluate the impact of the meta-learning approach on the quality of the federal dynamics model in the server. 

\begin{figure*}[ht]
    \centering
	\begin{subfigure}[ht]{0.48\linewidth}
        \centerline{\includegraphics[width=\linewidth]{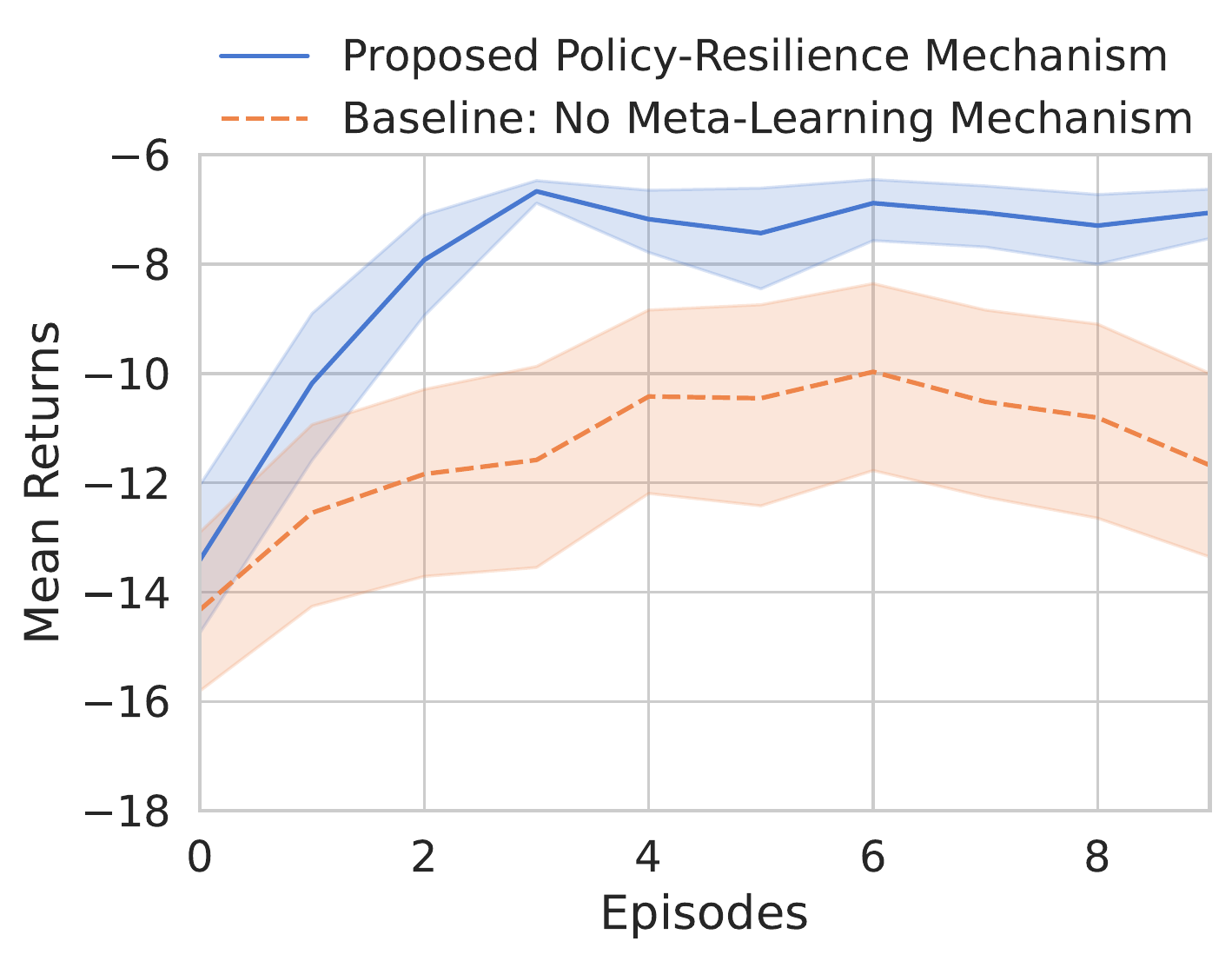}}
        \caption{poisoning proportion = 50\%}
        \label{fig:prop_50}
	\end{subfigure}
	\begin{subfigure}[ht]{0.48\linewidth}
        \centerline{\includegraphics[width=\linewidth]{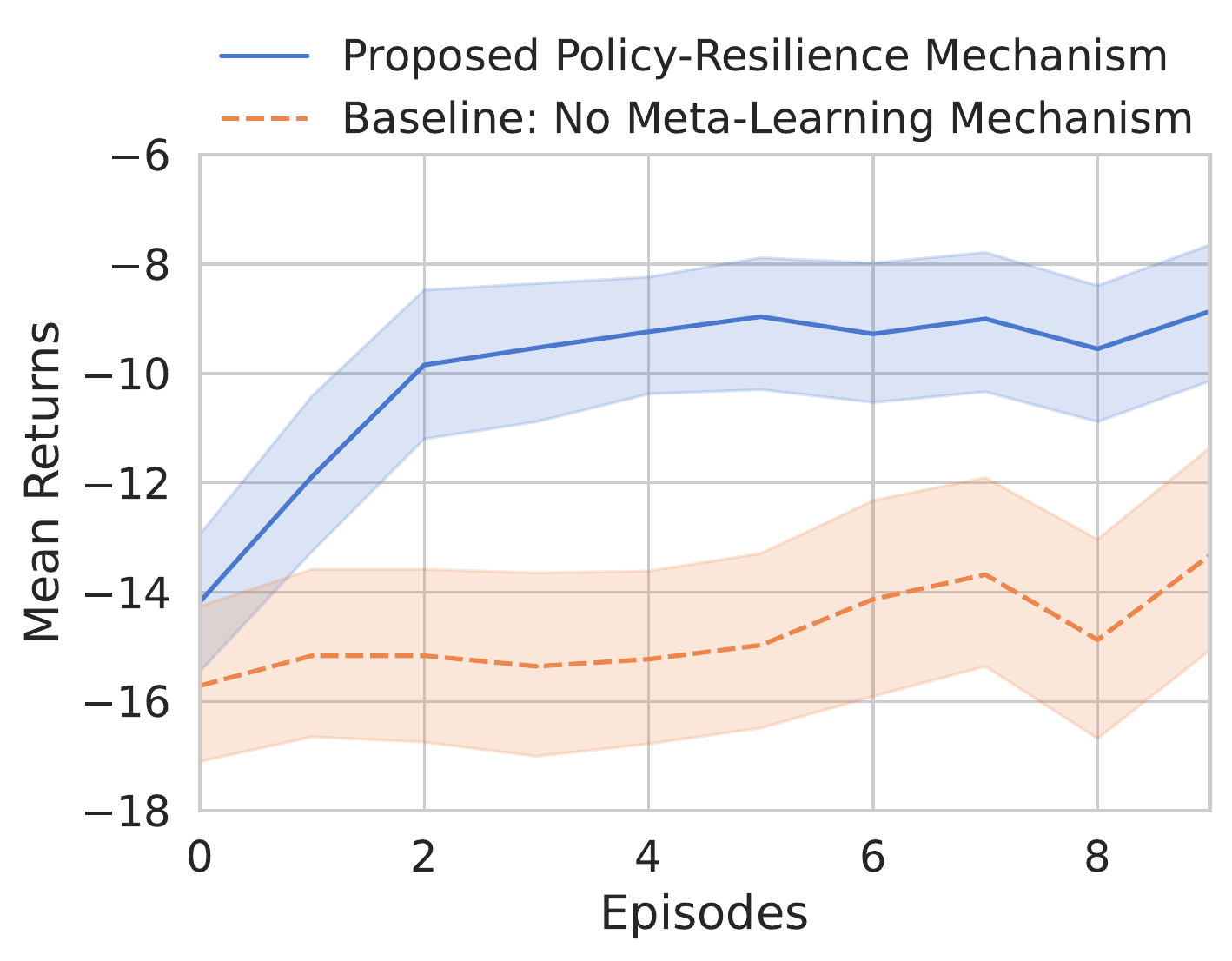}}
        \caption{poisoning proportion = 100\%}
        \label{fig:prop_100}
    \end{subfigure}
	\caption{Comparison of policy resilience under various proportions of poisoned clients.}
	\label{fig-cmp-prop}
\end{figure*}

% 图片-3 描述
First, Figure \ref{fig:3D_grid} shows that both the baseline {\em No Meta-Learning Mechanism} and our policy-resilience mechanism achieve good performances in recovering the policy.
It means that, when only one RL system has been attacked at the preparation stage, the both approaches lead to effective knowledge extraction (i.e., the learning of a server dynamics model).
% 图片-4-50% 描述
Then, we observe that, as the proportion of poisoned client RL systems increases, the effect of meta-learning approach becomes more apparent.
For example, as shown in Figure \ref{fig:prop_50}, when the proportion of poisoned clients reaches $50\%$, our policy-resilience mechanism still succeeds in recovering the agent's poisoned policy, whereas the performance of baseline {\em No Meta-Learning Mechanism} is somewhat decreased. 
% 解释
Here, this decreased performance can be attributed to the fact that the information collected from those poisoned RL systems has negatively impacted the learning of the server dynamics model.
In contrast, the meta-learning approach is not a simple combination of the federated dynamics model. 
It considers these poisoned RL systems as task instances and learns to how to efficiently learn the dynamics model of each task instance (i.e., namely "learn to learn"). 
Based on the meta-learning process (see Eq. \ref{eq:preparation-objective}), the server dynamics model is a parameterized function which is optimized to grasp the dynamics feature from a small set of training data.

% 图片-4-100% 描述
Furthermore, Figure \ref{fig:prop_100} shows the policy-resilience performances when all client RL systems are poisoned at the preparation stage.
Our policy-resilience mechanism is still effective in recovering the poisoned policy despite the slight reduction in efficiency and performance, while the baseline totally fails.
% 解释
This failure of the baseline indicates that the server dynamics model, which is learned using Federated Averaging Learning \cite{mcmahan2017communication}, cannot accurately model the natural deployment because the dynamics model is completely learned based on information generated by corrupted environments.
In contrast, in our policy-resilience mechanism, the slight performance reduction can be attributed to the fact that the deployment environment is not included in the task sets at the preparation stage.
In spite of this, the server dynamics model has been programmed with a learning ability, which enables it to model the dynamics of a task instance even if the task has not been displayed during preparation. 
In conclusion, the inclusion of the meta learning in our policy-resilience mechanism is crucial.

%% --------------------------------------------- %%
\subsection{Continuous State Domains} %% ------- %%
%% --------------------------------------------- %%

In this part, we evaluate our policy-resilience mechanism on both model-free and model-based RL agents in continuous state domains. 

\subsubsection{Experiment Settings}

\begin{figure}[ht]
    \centering
    \includegraphics[width=0.9\linewidth]{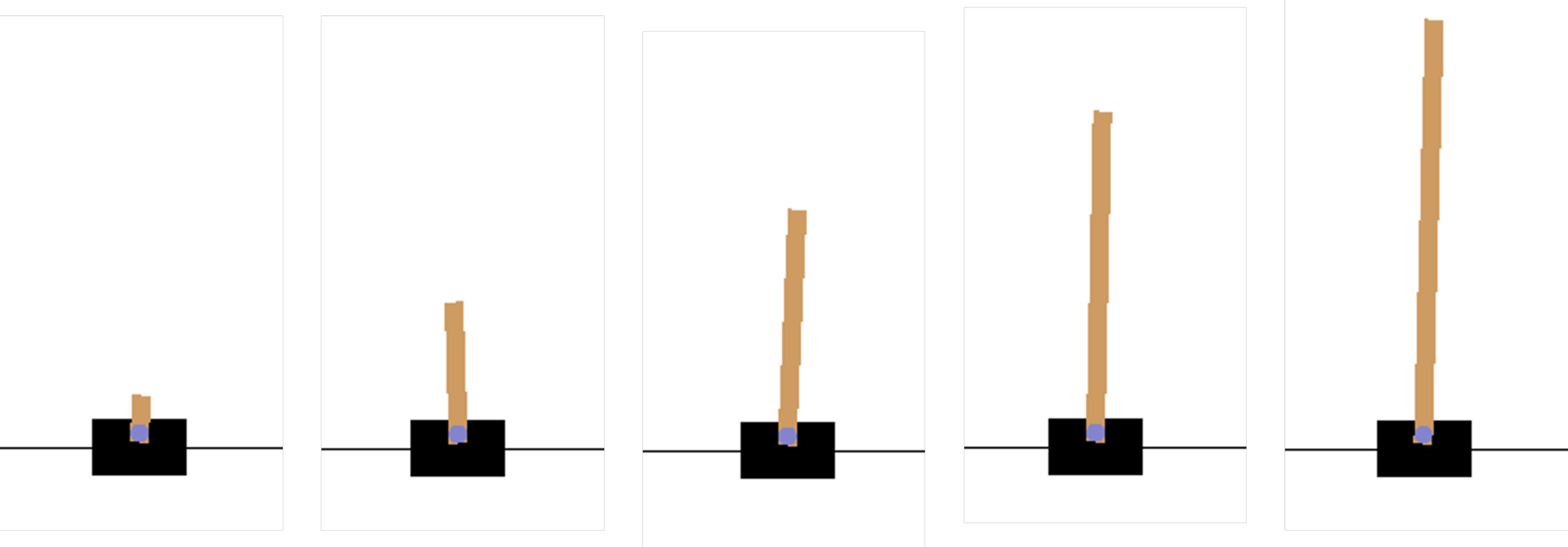}
    \caption{Illustration of Cartpole environment.}
    \label{fig:env_cartpole}
\end{figure}

\paragraph{Environments.}
The environment-poisoning attack, DBB-EPA, targets an RL agent performing a Cartpole task as depicted in Figure \ref{fig:env_cartpole}.
It is the agent's goal to balance the cartpole by exerting force in either the left or right directions on it.
When a force is applied, the angle and velocity of the pole are affected by the length of the pole.
As a result, the pole length can be viewed as the environment hyper-parameter which could be maliciously altered by an attacker.

\paragraph{Baselines.}
In this experiment, we design two baselines to evaluate our proposed policy-resilience mechanism.
\begin{itemize}
    \item Direct Policy Implementation without Resilience:
    To evaluate the performance of our policy-resilience mechanism, we use the policy learned in the poisoned training environments as the baseline.
    In other words, the poisoned policy will be implemented directly in the natural deployment environment.
    Such a baseline is termed as {\em Baseline: no policy resilience}.
    \item Policy Resilience without Knowledge Sharing: 
    To evaluate the design of federated framework associated with a meta-learning mechanism, we design a baseline policy-resilience procedure in which policy is recovered on the basis of its local dynamics model without the shared knowledge.
    We refer to this baseline as {\em Baseline: isolated policy resilience}.
\end{itemize}

\subsubsection{Experiment Results}

In light of the fact that an agent could adopt either model-free or model-based RL learning algorithms, we evaluate our policy-resilience mechanism on these two types of RL agents, respectively. 

\paragraph{Model-free RL Agents.}

%% 实验设置
This experiment involves 10 client RL systems within the policy-resilience framework, all of whose environments have been manipulated by the attacker during preparation. 
We simplify the attack implementation by manipulating the pole length only once at the beginning of the agent's learning process.
Thus, the RL agent learns its policy in the Cartpole environment where the pole length is manipulated as $0.5$, but will use the learned policy to control a car in which the pole length is $0.1 \sim 0.9$ (see Figure \ref{fig:env_cartpole}).
The agent learns the policy using model-free RL algorithms (e.g., DDPG) and recover it using MVE based on imagined trajectories generated by the learned dynamics model. 

%% 实验结果描述
Figure \ref{fig:cartpole_mve} illustrates the final performance of policy recovery within ten episodes, where the maximum value of scores is set as $500$. 
As shown in the figure, our proposed policy-resilience mechanism achieves nearly the maximum value of scores in restoring policy performance, which presents a significant improvement over two baselines. 
Additionally, we observe that {\em Baseline:isolated policy resilience} performs worse than {\em Baseline:no policy resilience}, which indicates that recovered policies are adversely affected by locally fine-tuned dynamics models that do not accurately reflect deployment environments.
We conclude from the comparison results that the design of our policy-resilience mechanism is capable of extracting and sharing critical knowledge about the environments, thus facilitating the development of an accurate dynamic model that generates imagined paths for effectively recovering the poisoned policy.

% \begin{figure}[ht]
%     \centering
%     \includegraphics[width=0.7\linewidth]{Figures/resilience_mve.pdf}
%     \caption{Comparison of policy-resilience performance on an \textit{model-free} RL agent performing the Carpole task.}
%     \label{fig:cartpole_mve}
% \end{figure}

\begin{figure}[ht]
    \centering
	\begin{minipage}[ht]{0.48\linewidth}
        \includegraphics[width=\linewidth]{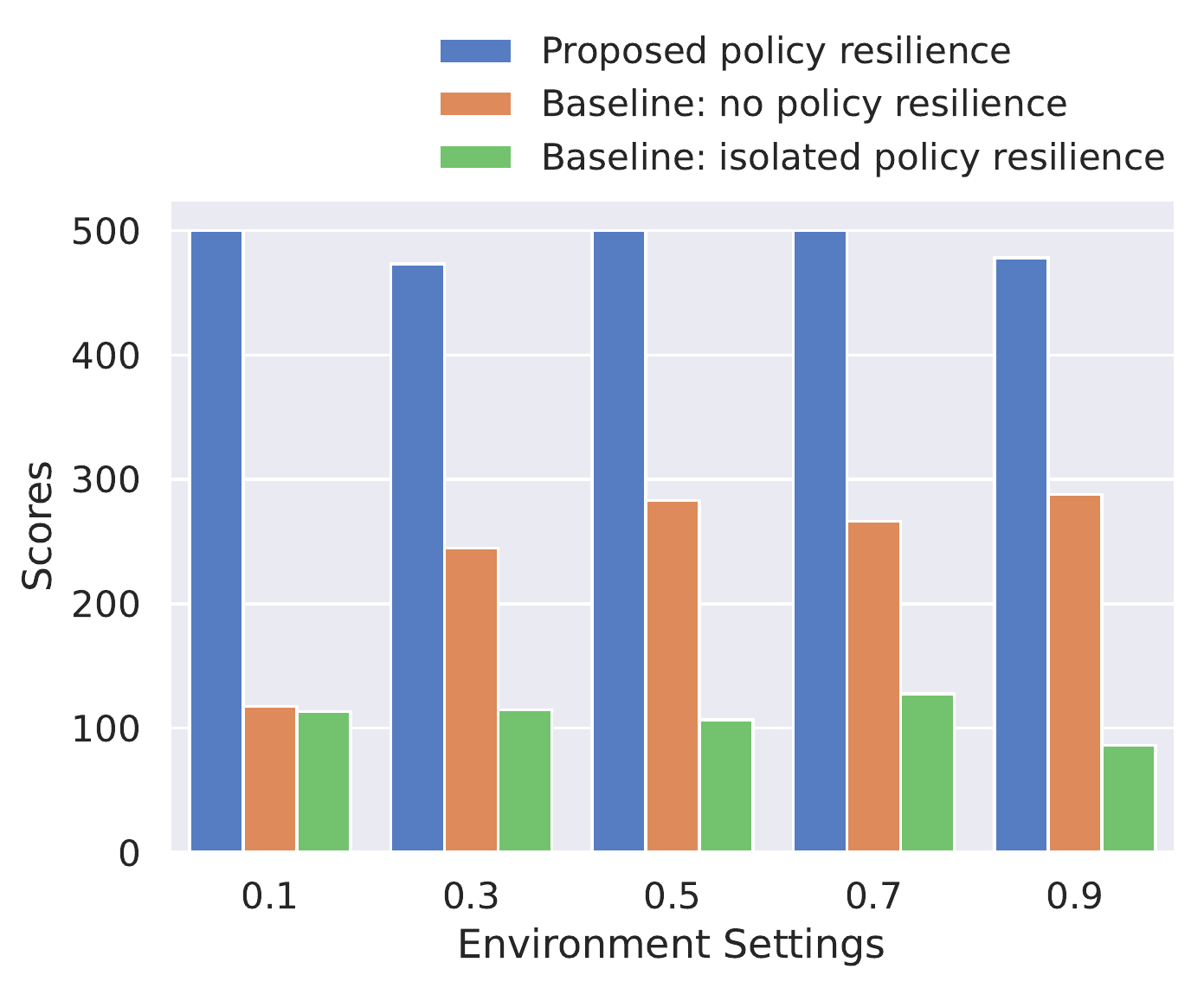}
        \caption{Comparison of policy-resilience performance on an \textit{model-free} RL agent performing the Carpole task.}
        \label{fig:cartpole_mve}
	\end{minipage}
	\begin{minipage}[ht]{0.48\linewidth}
        \includegraphics[width=\linewidth]{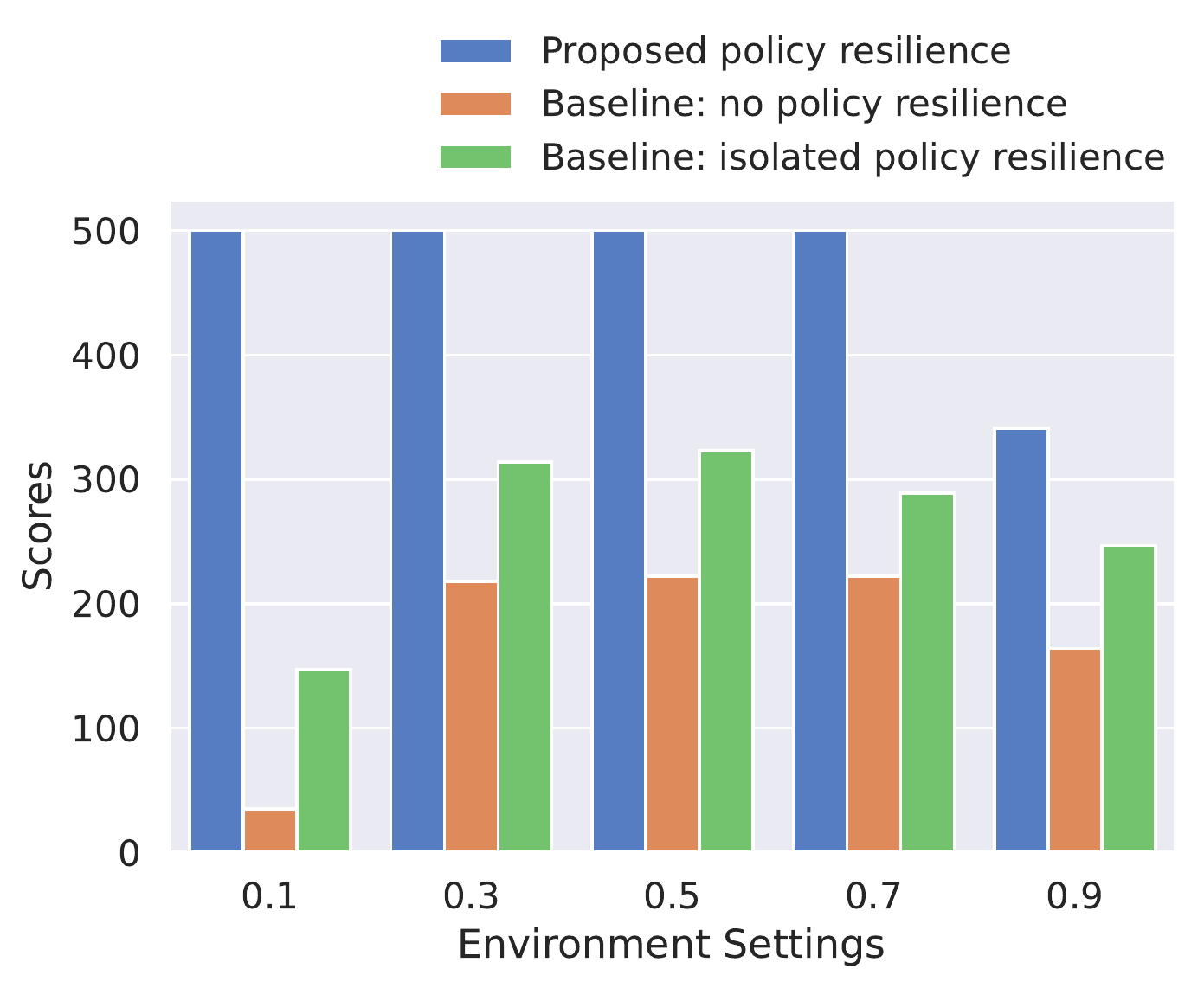}
        \caption{Comparison of policy-resilience performance on an \textit{model-based} RL agent performing the Carpole task.}
        \label{fig:cartpole_mpc}
    \end{minipage}
\end{figure}

\paragraph{Model-based RL Agents.}

We have also performed similar experiments on federated model-based RL agents that are trained to control the Cartpole using MPC algorithms.
Except for this difference, all other settings are the same as those used with model-free RL agents. 
Figure \ref{fig:cartpole_mpc} illustrates the performance of policy recovery in deployment environments where pole lengths are set at $0.1 \sim 0.9$ (see Figure \ref{fig:env_cartpole}).
As a result of adopting our policy-resilience mechanism, the agent achieves much higher scores than baselines in all deployment environments, which indicates the success in recovering poisoned policies.
Additionally, {\em Baseline:isolated policy resilience} performs better than {\em Baseline:no policy resilience} because of its locally updated dynamics model, however, it still performs poorly than our proposed approach particularly when training environments and deployment environments apparently varies in hyper-parameters.
Consequently, an isolated update of the dynamics model cannot provide an accurate representation of the environment, resulting in somewhat limited policy recovery performance.
In contrast, our policy-resilience mechanism encourages the sharing of critical environmental knowledge that facilitates the efficient development of an accurate dynamics model, which enables poisoned policies to be effectively recovered. 

% \begin{figure}[ht]
%     \centering
%     \includegraphics[width=0.7\linewidth]{Figures/resilience_mpc.pdf}
%     \caption{Comparison of policy-resilience performance on an \textit{model-based} RL agent performing the Carpole task.}
%     \label{fig:cartpole_mpc}
% \end{figure}

% 对比讨论
We also observe a significant difference between {\em Baseline:isolated policy resilience} in Figures \ref{fig:cartpole_mpc} and \ref{fig:cartpole_mve}, namely that locally updated dynamics models have a positive impact on policy recovery for model-based agents whereas they have a negative impact on policy recovery for model-free agents.
This difference is explained by the fact that the model-free agent that uses MVE to recover its policy is more sensitive to the inaccuracy of the local dynamics model.
As a result, this observation indirectly shows that the dynamics model learned by our policy-resilience mechanism is sufficiently accurate for the model-free agent to achieve the policy recovery objective.

%% ===========================================================================
\section{Further Discussion}
%% discussion

In this section, we provide detailed discussions of the assumptions and limitations of this work, including three points: the security issues of federated learning, the application scope of policy-resilience design, and the learning effectiveness of dynamics models.

\paragraph{Security Issues of Federated Framework.}
Throughout this work, we design the policy-resilience mechanism in a federated manner, because federated learning is a privacy-aware paradigm of model training.
However, recent studies have indicated that FL does not always provide sufficient protection for privacy and robustness \cite{lyu2020privacy}.
For example, it is possible that 
(1) an adversary server may attempt to gather sensitive information from individual updates over time, interfere with training procedures, or restrict participants' views of global parameters; 
(2) malicious participants may have the potential to disrupt the process of aggregating global parameters, poison the global model, or infer sensitive information about other participants.
Accordingly, the security issue of federated learning is a significant research topic that deserves further exploration, however, it is not the focus of our study.
As a result, our policy-resilience mechanism is built on the assumption that the security of federated frameworks is guaranteed.

\paragraph{Applicability of Policy-Resilience Design.}
The applicability of our policy-resilience mechanism is discussed from two viewpoints: its potential limitations and its extended scope. 

Our policy-resilience framework is constructed from a set of independent RL systems, which can result in restricted application if only one agent is present.
Nevertheless, our proposed mitigation framework is applicable in many RL applications, such as robot cloud and robots \cite{liu2019lifelong}, a state grid corporation and building grids \cite{lee2020federated}, and a transport authority and autonomous vehicles \cite{liang2019federated}.
Furthermore, as the Internet of Things (IoT) \cite{li2015internet} and federated learning \cite{mcmahan2017communication} advance, more isolated systems will be connected together for model learning or security protection, resulting in a wider application scope for our policy-resilience design. 

We further claim that our policy-resilience approach can be applied to address policy adaptation problems, in which a learned RL policy is competent but specialized so that it will be ineffective in an unknown deployment environment.
This problem is possible to be addressed by a classical meta-learning approach \cite{zintgraf2019varibad,finn17maml}.
However, the classical meta-learning approach assumes that the RL agent can access multiple environments throughout the training process, which is commonly impossible due to physical or security issues in real-world applications.
For instance, when training a self-driving car in the tropics (e.g. Singapore), it is difficult to reproduce an icy road and snowy surroundings as a training environment. Therefore, this car would be at a loss in colder regions (e.g. Moscow) where such road conditions are natural.
In this example, the physical issue prevents the RL agent from accessing various training environments.
In light of this, agents, which can only learn their policy in a single training environment, should be able to quickly comprehend unknown deployment environments and accordingly improve policy performance.
Our proposed policy-resilience mechanism, which incorporates a federated framework with a meta-learning mechanism, can fulfill this goal.

%% 关于 dynamics model 的学习难度
\paragraph{Learning Efficacy of Dynamics Model.}
In our work, the quality of the server dynamics model plays an influential role on the performance of policy recovery.
As a result, the learning efficiency of the dynamics model is a key point in our policy-resilience mechanism.
Dynamic modeling, however, may become more challenging as the environment becomes more complex.
This may pose challenges to policy-resilience performance, which is a potential concern in our work.
Nevertheless, this concern is expected to be addressed since a number of methods \cite{chua2018deep, antonoglou2021planning, schrittwieser2021online} have been proposed to accurately represent the dynamics of the environment and these methods can be incorporated into our policy-resilience mechanisms.

\end{document}